\pdfoutput=1
\documentclass[11pt]{article}

\usepackage[final]{acl}

\usepackage{times}
\usepackage{latexsym}
\usepackage{amsmath}

\usepackage[T1]{fontenc}
\usepackage[utf8]{inputenc}

\usepackage{microtype}

\usepackage{inconsolata}

\usepackage{booktabs}
\usepackage{amssymb}
\usepackage{graphicx}
\usepackage{ragged2e} 
\usepackage{float}
\usepackage{xcolor}
\usepackage{mdframed}
\usepackage{array}

\usepackage{array,booktabs}

\newcolumntype {+}{ >{\global\let\currentrowstyle\relax}}
\newcolumntype {^}{ >{\currentrowstyle }}

\newcommand{\mysubsection}[1]{\vspace{0.3em}\noindent\textbf{#1}}

\title{The Impact of Annotator Personas on LLM Behavior Across the Perspectivism Spectrum}

\author{\textbf{Olufunke O. Sarumi\textsuperscript{1}},
    \textbf{Charles Welch\textsuperscript{2}},
    \textbf{Daniel Braun\textsuperscript{1}},
    \textbf{Jörg Schlötterer\textsuperscript{1,3}}
    \\ 
    \textsuperscript{1}University of Marburg, \textsuperscript{2}McMaster University,
    \textsuperscript{3}University of Mannheim\\
    \small\{sarumio,daniel.braun,joerg.schloetterer\}@uni-marburg.de\textsuperscript{1}, cwelch@mcmaster.ca\textsuperscript{2}
}

\begin{document}
\maketitle
\begin{abstract}
In this work, we explore the capability of Large Language Models (LLMs) to annotate hate speech and abusiveness while considering predefined annotator personas within the strong-to-weak data perspectivism spectra. We evaluated LLM-generated annotations against existing annotator modeling techniques for perspective modeling. Our findings show that LLMs selectively use demographic attributes from the personas. We identified prototypical annotators, with persona features that show varying degrees of alignment with the original human annotators. Within the data perspectivism paradigm, annotator modeling techniques that do not explicitly rely on annotator information performed better under weak data perspectivism compared to both strong data perspectivism and human annotations, suggesting LLM-generated views tend towards aggregation despite subjective prompting. However, for more personalized datasets tailored to strong perspectivism, the performance of LLM annotator modeling approached, but did not exceed, human annotators.

\end{abstract}

\section{Introduction}

Perspectivism in Natural Language Processing (NLP) aims to preserve the spectrum of opinions held by annotators in corpora \citep{Cabitza_Campagner_Basile_2023}. Dataset annotation for this purpose often uses a descriptive paradigm \citep{rottger-etal-2022-two}, involving minimal instructions and multiple annotators providing labels for every corpus sentence to capture diverse viewpoints. The number of annotators involved can range significantly, from a minimum of 2 to 2500 or more \citep{plepi-etal-2022-unifying, frenda2024perspectivist}.

Most traditional approaches aggregate labels to obtain a single majority label \citep{davani-etal-2022-dealing, aroyo2015truth}, which is commonly used for training models. However, the perspectivist approach argues that critical information is lost when labels are aggregated. More importantly, the opinions of the minority, which may represent a significant population, are undermined, leading to under-representation and overshadowing of nuances inherent in the dataset. This is crucial because people's views and opinions are indeed shaped by different socio-cultural, demographic, economic, and experiential backgrounds \citep{akhtar2021opinionsmatterperspectiveawaremodels, almanea-poesio:2022:LREC, demszky2020goemotions, kennedy2022gab}. These factors impact how individuals perceive, interpret, and respond to various topics, making it unrealistic to assume everyone shares similar views on the same subject. Recognizing and reflecting opinion differences in our models is therefore important for developing socially aware NLP systems, treating disagreements not as errors but as distinct perspectives. To address this, models have been developed that can learn from such disaggregated labels \citep{leonardelli-etal-2023-semeval,sullivan-etal-2023-university,vitsakis-etal-2023-ilab,garcia-diaz-etal-2023-umuteam-semeval,cui-2023-xiacui,xu2024leveragingannotatordisagreementtext}.

Furthermore, while some disagreements stem from different perspectives, other factors also cause disagreement in data annotations, including temporal factors, annotator inconsistencies, uncertainty, ambiguities, lack of task understanding, or a perfunctory approach to annotation \citep{fleisig-etal-2024-perspectivist}. When modeling perspectives obtained from subjective tasks, these perspectives are often mixed with noise and errors, raising the question of whether true perspectives or merely annotator inconsistencies have been modeled. Some literatures have quantified these uncertainties to a minimal extent \citep{klemen-robnik-sikonja-2022-ulfri,davani-etal-2022-dealing}.

In this work, we aimed to investigate how existing annotator modeling techniques would behave when trained on deterministic LLM-generated annotations, in contrast to earlier work that explored modeling individual human annotators' perspectives using disaggregated labels. We generated new annotations for the HS-Brexit and ConvAbuse datasets using Llama2-13B, guided by persona-based prompting derived from annotator information provided by the original authors.

In generating these annotations, we implemented two perspectivism approaches: \textit{strong} and \textit{weak} data perspectivism. Weak perspectivism, also known as reduced perspective, involves considering multiple labels which are ultimately aggregated into one, representing a group opinion. Strong perspectivism, by contrast, utilizes and retains all distinct labels from training through evaluation \citep{Cabitza_Campagner_Basile_2023, frenda2024perspectivist}.

Our findings show that LLMs struggle to generate responses as diverse as humans, even with diverse personas. They still partially align with human annotations but tend to pick up only selected persona features. Furthermore, we identified latent annotation prototypes shared by multiple human annotators. These alignment patterns vary across datasets and perspectivism strategies: for instance, HS-Brexit with contrasting demographic attributes shows stronger alignment with human annotations under weak perspectivism, whereas ConvAbuse demonstrates closer alignment with human annotations when strong data perspectivism is used, involving highly personalized and overlapping persona features.

\section{Related Work}
The first part of this section addresses how Large Language Models (LLMs) have been used to generate different perspectives and their ability to adopt an assigned persona. It also highlights the lack of connection between perspectivism, based on defined personas and annotations in subjective tasks. The second part focuses on the use of LLMs as annotators, examining their ability to generate discrete multiple labels, identifying the lack of persona-based labeling, and replicating human annotation behavior to enable alignment with human annotations.

\subsection{LLMs in Perspectivism and Adopting Personas}
LLMs have been explored for their ability to simulate diverse human perspectives. Subjective tasks often involve annotators with different backgrounds, leading to divergent opinions which often reflect demographic variation, different and substantial opinions, these make label aggregation inadequate \citep{rottger-etal-2022-two}.
Some works argue that LLMs naturally contain persona traits, as they are trained in large corpora, often culled from social networks that contain crowd-sourced data rich with diverse viewpoints \citep{hu2024quantifyingpersonaeffectllm, vitsakis-etal-2023-ilab}. For example, \citet{hayati-etal-2024-far} showed that it is possible to generate multiple perspectives from LLMs and quantify the maximum number of perspectives derivable from an LLM. However, the influence of persona prompting remains debated and the influence of specific persona traits remains underexplored \citep {beck-etal-2024-sensitivity, sun2025sociodemographicpromptingeffectiveapproach}. \citet{hu2024quantifyingpersonaeffectllm} suggests that personas have minimal effect on LLM outputs, whereas a psycholinguistic research found that LLMs can generate human-like outputs, even surpassing humans in turing experiments, yet exhibit unnaturally high accuracy that is not possible within human populations \citep{aher2023usinglargelanguagemodels}. Furthermore, \citet{wang2024largelanguagemodelsreplace} found that LLMs risk homogenizing or misrepresenting marginalized identity groups, particularly when asked to simulate them. These challenges highlight the difficulty in separating the LLM's inherent persona from externally applied persona prompts.
Despite this, prompting LLMs with well-defined personas, particularly those grounded in demographic traits from existing datasets, offers a practical way to examine how perspective alignment occurs between machines and humans. However, small variations in prompt configurations can lead to large differences in output, complicating reproducibility and fairness evaluations.

\subsection{LLM Annotations and Label Generation}
Beyond simulating perspectives, LLMs are being explored as direct substitutes for human annotators \citep{ivey2024realroboticassessingllms,bavaresco2024llmsinsteadhumanjudges}, especially in settings where collecting human annotations is expensive or slow \citep{huang-etal-2023-incorporating, gligorić2024unconfidentllmannotationsused}. Recent studies have examined the ability of LLMs to generate discrete labels for classification tasks, often using crowd-sourced datasets as benchmarks \citep{pavlovic2024effectivenessllmsannotatorscomparative, gilardi2023chatgpt}.
\citet{gilardi2023chatgpt} found that LLMs outperformed crowd-sourced workers in certain annotation tasks, while \citet{pavlovic2024understandingeffecttemperaturealignment} demonstrated that adjusting temperature values can control LLM behavior to better simulate annotation disagreement or consistency. These findings suggest that LLMs can be tuned to exhibit behavior similar to individual or aggregated human annotators.
LLMs have also been deployed in replicating prior annotation experiments. For example, \citet{pavlovic2024effectivenessllmsannotatorscomparative} replicated a Learning With Disagreement task \citep{leonardelli-etal-2023-semeval} using GPT-3 but did not incorporate the demographic background of annotators, limiting their insight into perspective-specific agreement.
While many experiments rely on LLMs generating explanations or engaging in dialogue-based tasks, fewer works have explored their ability to produce discrete, disaggregated annotations comparable to crowdsourced annotators. Likewise, existing annotator modeling techniques are yet to be fully evaluated on annotations generated by LLMs. The impact of LLM annotations and predefined personas on existing annotator modeling approaches remains unexplored and is a key area we address in our study.

\section{Dataset}
We used two datasets from the SemEval-2023 task on learning with disagreements ~\citep{leonardelli-etal-2023-semeval} and used Llama2-13B to generate annotations for weak and strong data perspectivism variants resulting in six (6) datasets. Strong perspectivism used prompts tailored to individual persona descriptions, while weak perspectivism used group descriptions to simulate aggregated viewpoints; however, the persona descriptions in each variant were limited to the demographic information and features provided in the original work. All datasets use binary labels for classification. Original dataset statistics are presented in Table \ref{tab:original_dataset}. 

\begin{table*}[t]
\small
\centering
\begin{tabular}{lccccc}
\toprule
& \#A & \#I & N & A/I & K-$\alpha$ \\
\midrule
HS-Brexit & 6 & 1,120 & 1,120.00$\pm$ 0.00 & 6.00$\pm$ 0.00 & 0.35\\
ConvAbuse & 8 & 4,050 & 1,521.00$\pm$ 206.91 & 3.00$\pm$ 0.88 & 0.65 \\
\bottomrule
\end{tabular}
\caption{Original Dataset Statistics by Human Annotators. \#A: number of annotators, \#I: number of total instances, N: number of annotations per annotator, A/I: annotations per instance, K-$\alpha$: Krippendorff’s alpha agreement.}
\label{tab:original_dataset}
\end{table*}

\mysubsection{HS-Brexit}
The Hate Speech Brexit (HS-Brexit) dataset \citep{Akhtar2021WhoseOM} comprises 1,120 tweets concerning Brexit and immigration, annotated for hate speech, aggressiveness, and offensiveness. This dataset features annotations from two distinct groups of three individuals: a target group of Muslims and first- or second-generation immigrants to the UK (also classified as migrants in the original study) and a control group of researchers with a Western background making six annotators in all.

\mysubsection{ConvAbuse}
 The Conversational Abuse (ConvAbuse) dataset, as described by \citet{cercas-curry-etal-2021-convabuse}, comprises roughly 4,000 English dialogues between users and two conversational agents. These user conversations were labeled by a minimum of three gender studies experts, using a hierarchical annotation system that included categories for presence, severity, and directness of abuse. We binarized the annotations into two classes, 0 and 1. The ConvAbuse dataset is characterised by eight $(8)$ annotators, each providing a significant number of annotations. Also, not all the 8 annotators labeled every instance contrary to the HS Brexit, but each annotator has annotations.

\section{Methodology}
Firstly, we explore the ability of Llama2-13B to generate discrete binary annotations on the datasets, using defined personas. Secondly, we modeled these personas with existing annotator modeling techniques. 
\subsection{Annotation Generation}
\begin{figure*}
    \centering
    \includegraphics[width=\linewidth, clip]{ 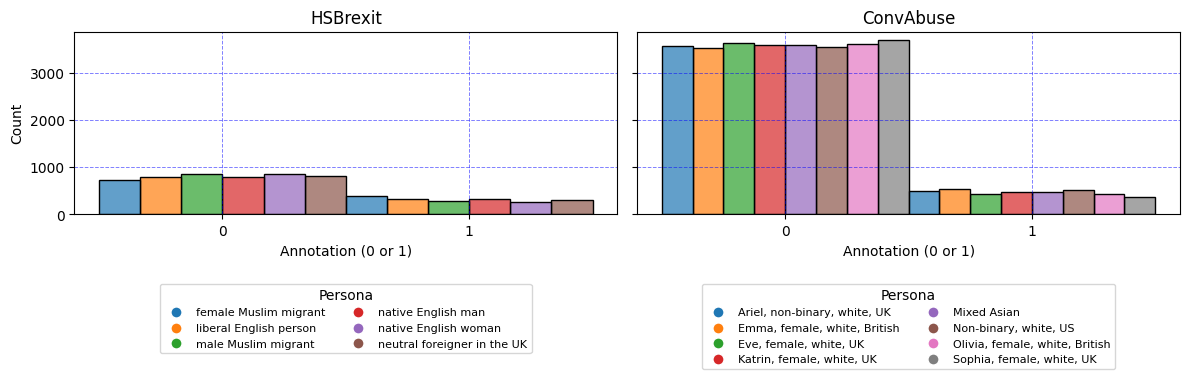}
    \caption{Distribution of Annotations for ConvAbuse and HS-Brexit (Strong Perspective) by Persona}
    \label{fig:conv_strong}
\end{figure*}

For the strong perspectivism variant of the datasets, we prompt Llama2-13B with each text in the original corpus. We extended the dataset with the generated annotations for each corresponding persona, maintaining the original structure of the dataset from the SemEval-2023 task. The \textit{strong} variant uses specific individual descriptions for each persona as seen in Figure \ref{fig:conv_strong}. In the original ConvAbuse dataset, not all annotators annotated all instances, but in the LLM version, all eight annotators were represented in all instances. We generate annotations at temperatures: 0, 0.1, 0.2, 0.5 and 0.8, for each perspectives. We used the demographic description presented in the original work as guide for our persona features. 
\begin{table}[htbp]
	\centering
	\begin{mdframed}[
		backgroundcolor=black!5,
		linewidth=1.2pt,
		innerleftmargin=2pt,
		innerrightmargin=2pt,
		innertopmargin=2pt,
		innerbottommargin=2pt,
		skipabove=1em,
		skipbelow=1em,
		linecolor=black,
		innermargin=0pt,
		leftmargin=0.05\linewidth,
		rightmargin=0.05\linewidth
	]
		\small
		\textit{annotate this statement regarding Brexit.
			as a group of (persona) in the UK or Europe, you must assign a label to the statement Use '0' if the statement does not express hate towards Brexit, and '1’ if it does express hate towards Brexit. Provide only the number as your response without any additional text or explanation for example "\#\#\#\#Annotator:"\#\#\#\#Annotator:0" or "\#\#\#\#Annotator:1"
		}
	\end{mdframed}
	\caption{\small An example prompt for weak perspectivism in HS-Brexit}
	\label{tab:prompt_weakhs}
\end{table}
In \textit{weak} perspectivism, we followed the same approach. Figures \ref{fig:alignmentconv} and \ref{fig:alignmenthsb} show persona descriptions and Table \ref{tab:prompt_weakhs} shows a sample of the prompt used. The prompt and personas are fully described in the Appendices \ref{sec:prompt} and \ref{sec:persona}, respectively. Also in Table \ref{tab:llama2_dataset}, we show a summary of the data statistics and the variance observed in the inter-annotator agreement K-$\alpha$ as temperature increases.

\begin{table*}[t]
\small
\centering
\begin{tabular}{lcclccc}
\toprule
%\multicolumn{7}{c}{\textbf{LLAMA2- 13B Dataset Statistics for Strong and Weak Perspectives}} \\
%\midrule
& \#A & \#I & N & A/I & K-$\alpha$ (Strong) & K-$\alpha$ (Weak) \\
\midrule
HS-Brexit & 6 & 1,120 & 1,120.00$\pm$ 0.00 & 6.00$\pm$ 0.00 & 0.58 -- 0.81 (T=0.8 -- 0) & 0.55 -- 0.75 (T=0.8 -- 0) \\
ConvAbuse &  8 & 4,050 & 4,050.00$\pm$ 0.00 & 8.00$\pm$ 0.00 & 0.60 -- 0.91 (T=0.8 -- 0) & 0.62 -- 0.93 (T=0.8 -- 0) \\
\bottomrule
\end{tabular}
\caption{LLAMA2 Dataset Statistics. \#A: number of annotators, \#I: number of total instances, N: number of annotations per annotator, A/I: annotations per instance, K-$\alpha$: Krippendorff’s alpha agreement (T=temperature range). The K-$\alpha$ values are presented as a range from temperature 0.8 to 0, that is agreement decreases as temperature increases.}
\label{tab:llama2_dataset}
\end{table*}

\subsection{Annotator Modeling}
We trained existing annotator models \cite{oluyemi-etal-2024-corpus, davani-etal-2022-dealing} using the LLM-generated labels, following a classification pipeline originally used with the human-annotated corpus.
 We replicated these annotator modeling techniques—User Token, Composite Embedding, Composite+User Token Embedding, and Multi-task to model perspectives by modeling annotators, and we also added a text-only implementation without annotator information with SBERT. These techniques used annotator IDs and text, with 6 annotations per instance for the HSBrexit and 8 annotations per instance for the ConvAbuse to capture the \textit{persona} perspectives using the labels obtained from the generations at all temperatures but used the best scores (generally between temperatures 0 and 0.1) in our results and analysis. The annotator ID represents each unique annotator(persona), encoded as embeddings. Each technique uses a different method to generate encodings used to uniquely model the personas.
Finally, we compared the performance of these annotator modeling techniques on LLM-generated annotations and human annotations. 
\begin{table*}[h]
\small
\centering
\begin{tabular}{lccccc}
\toprule
Method & SBERT & User Token & Composite Embedding & Composite Embedding + User Token & Multi-Tasking \\
\midrule
\multicolumn{6}{c}{\textbf{Human-annotations }} \\
\midrule
HS-Brexit & 68.6 & 77.6 & 67.6 & 77.3  & 71.7 \\
ConvAbuse & 85.9 & 88.5 & 85.8  & 88.6 & 82.3 \\
\midrule
\multicolumn{6}{c}{\textbf{LLAMA2-13B strong perspectivism}} \\
\midrule
HS-Brexit & \textbf{72.2} & 69.4 & \textbf{71.8} & 71.2 & 65.1\\
ConvAbuse & 85.7 & 84.4 & 84.6 & 84.4 & 81.1 \\
\midrule
\multicolumn{6}{c}{\textbf{LLAMA2-13B weak perspectivism}} \\
\midrule
HS-Brexit & \textbf{73.2} & 72.2 & \textbf{72.4} & 71.7 & 62.0\\
ConvAbuse & 85.2 & 83.7 & 83.7 & 81.8 & 79.8 \\
\bottomrule
\end{tabular}
\caption{Model performance based on individual annotator and persona F1 scores. Results for human annotations was adapted from \citet{oluyemi-etal-2024-corpus}. We reported the best LLM results for temperatures 0 and 0.1.}
\label{tab:model}
\end{table*}

\section{Results}
Table \ref{tab:model} presents the F1 scores for the annotator modeling techniques evaluated on both the original and the synthetic datasets. Our analyses show some trends in the performance of these models. In existing results that used human annotations, overall performance was observed on the ConvAbuse dataset. The inter-annotator agreement measured by Krippendorff’s alpha was high for ConvAbuse and comparatively lower for the HSBrexit dataset. Interestingly, the Llama2-annotated versions showed significantly higher agreement levels than the original human annotations across all temperature settings, including at a high randomness level (Temperature = 0.8) as seen in Tables \ref{tab:original_dataset} and \ref{tab:llama2_dataset}.
 Prior research established that the effectiveness of annotator modeling techniques is largely dependent on the degree of agreement and the number of annotations per annotator \cite{oluyemi-etal-2024-corpus}. Specifically, the User-Token modeling approach performs best for datasets with low agreement, while the Composite Embedding + User Token method is optimal for datasets with high agreement. Both methods rely on an explicit naming system, using annotator IDs to individually predict the label outputs for each annotator.
 However, our results indicate that models without explicit annotator information outperformed others on the Llama2 persona-based datasets. For instance, SBERT, with no annotator information and Composite Embedding- an approach that did not use explicit naming convention (annotator ID) for modeling, both outperformed the best-performing models on HSBrexit and achieved comparable results on ConvAbuse. This suggests that the optimal annotator modeling techniques for human annotations may not be directly transferable or equally effective for data annotated through LLM personas.
\subsection{Strong vs Weak Data Perspectivism in Annotator Modeling}
As presented in Tables \ref{tab:strong_weak_convabuse} and \ref{tab:strong_weak_hsb} of Appendix \ref{sec:strong_weak}, we adapted the two versions of data perspectivism described by \cite{Cabitza_Campagner_Basile_2023} and evaluated the annotator modeling techniques on the datasets. The strong perspectivist approach, which used fine-grained persona profiles, generally produced higher performance that was more aligned with the results from human modeling for the ConvAbuse dataset at temperature 0.1. The weak perspectivism approach, characterized by contrasting group descriptions, showed improved performance over the human version in the HS-Brexit dataset across both strong and weak variants, with a greater improvement observed in the weak, group-based variant. However, this performance increase was exclusively observed in the Composite Embedding and SBERT models without explicit annotator information.

\begin{figure*}[htbp]
    \centering
    \begin{mdframed}[
        backgroundcolor=white,
        linewidth=0pt,
        innerleftmargin=1pt,
        innerrightmargin=1pt,
        innertopmargin=1pt,
        innerbottommargin=1pt,
        roundcorner=0pt,
        skipabove=0pt,
        skipbelow=0pt,
        leftmargin=0.05\linewidth,
        rightmargin=0.05\linewidth
    ]
        \centering
        \includegraphics[width=0.48\linewidth]{ 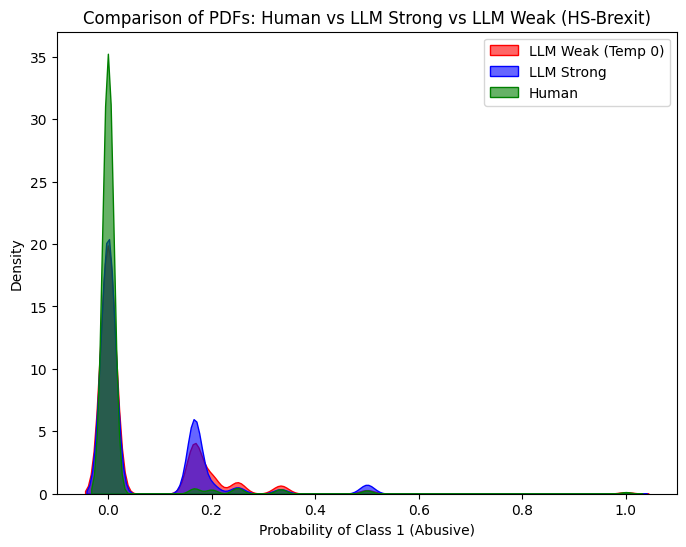}
        \hfill
        \includegraphics[width=0.48\linewidth]{ 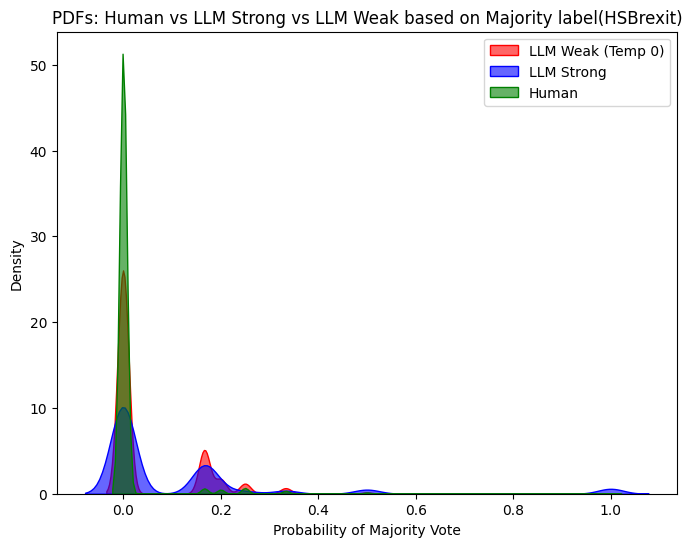}
    \end{mdframed}
    \caption{\small Figure showing the Probability Density Function illustrating Uncertainty in LLM annotations Vs Human in HSBrexit}
    \label{fig:pdfhs}
\end{figure*}

\subsection{Annotation Quality and Uncertainty}
 We analyzed the quality of annotations generated by Llama2-13B across a spectrum of temperature parameters. Even at high randomness with temperature set to 0.8, inter-annotator agreement remained high cf. Table \ref{tab:llama2_dataset}. The distribution of labels diverged significantly from that of the human annotators. To illustrate this, we compared the label distributions using Probability Density Functions (PDFs). The human annotations showed a sharp peak near class 0, indicating a highly consistent assignment of non-abusive class, despite disagreement, in the HS-Brexit dataset as seen in Figure \ref{fig:pdfhs}. In contrast, the PDF for the strong perspectivist variant of the LLM showed a slightly right-skewed peak between 0.1 and 0.2, suggesting that the LLM assigned marginally higher soft labels than human annotators. The weak perspectivist PDF was flatter and more dispersed, with a small density spike near a probability of 0.2, reflecting greater uncertainty and inconsistency in labeling. The PDFs for the ConvAbuse dataset is presented in the Appendix \ref{sec:pdf}.

\subsection{Prototypical \textit{Persona} Annotators and Human Alignment}

% \noindent
\textbf{Ablation 1}: Table \ref{tab:table_align} shows that annotator models trained on LLM annotations perform worse when tested on human labels, indicating a lack of alignment. The decline likely comes from the lack of corresponding match between LLM personas and the unknown individual human annotators.
\begin{figure*}[t!]
    \centering
    \begin{mdframed}[
        backgroundcolor=white,
        linewidth=0.2pt,
        innerleftmargin=0.2pt,
        innerrightmargin=0.2pt,
        innertopmargin=0.5pt,
        innerbottommargin=0.5pt,
        roundcorner=0pt,
        skipabove=0pt,
        skipbelow=0pt,
        leftmargin=0.05\linewidth,
        rightmargin=0.05\linewidth
    ]
        \centering
        \includegraphics[scale=0.5]{ 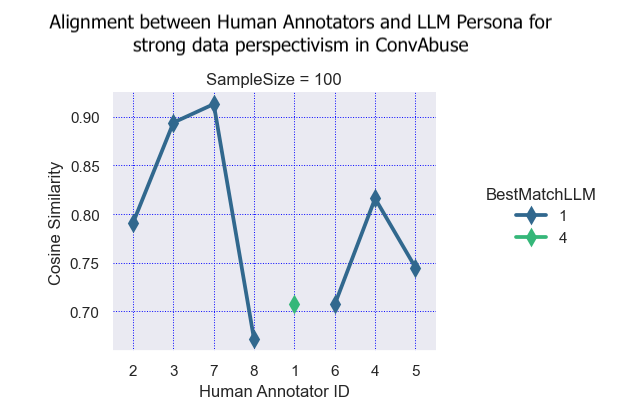}
        \hfill
        \includegraphics[scale=0.5]{ 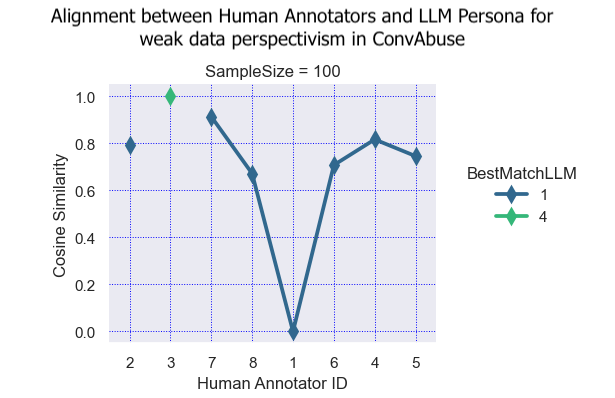}
        \vspace{1em}
        {\tiny
        \begin{tabular}{lll}
            \toprule
            Human/LLM ID & Persona (strong) & Persona (weak)\\
            \midrule
            1 & Olivia, a female and white British person & white British female people\\
            2 & Emma, a female and white British person & white British with non-binary gender orientation\\
            3 & Ariel, a white person from the United Kingdom with a non-binary gender orientation & non-binary gender people from the United States\\
            4 & Sophia, a female and white person from the United Kingdom & white female people from the United Kingdom\\
            5 & Katrin, a female and white person from the United Kingdom & white female from United States\\
            6 & Eve, a female and white person from the United Kingdom & mixed Asian with a non-binary gender orientation\\
            7 & mixed Asian person & mixed Asian female\\
            8 & a white person from the United States with a non-binary gender orientation & mixed Asian female\\
            \bottomrule
        \end{tabular}
        }
    \end{mdframed}
    \caption{\small Figure showing Prototypical LLM annotators and Alignment with Human Annotators in ConvAbuse}
    \label{fig:alignmentconv}
\end{figure*}
% \noindent

\begin{figure*}[t!]
    \centering
    \begin{mdframed}[
        backgroundcolor=white,
        linewidth=0.2pt,
        innerleftmargin=1pt,   % Slightly increased for better visual separation
        innerrightmargin=1pt,  % Slightly increased
        innertopmargin=4pt,    % Increased top/bottom padding for content
        innerbottommargin=4pt, % Increased
        roundcorner=0pt,
        skipabove=0pt,
        skipbelow=0pt,
        leftmargin=0\linewidth, % Reduce side margins to make the frame wider
        rightmargin=0\linewidth % Reduce side margins
    ]
        \centering
        \includegraphics[scale=0.5]{ 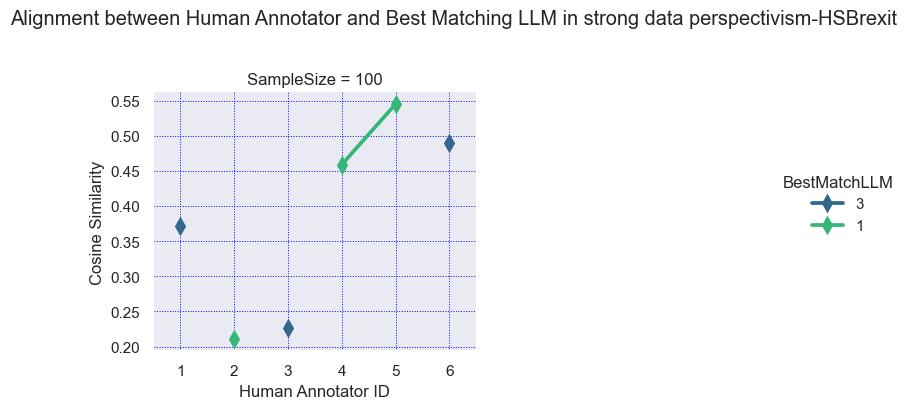}
        \hfill
        \includegraphics[scale=0.5]{ 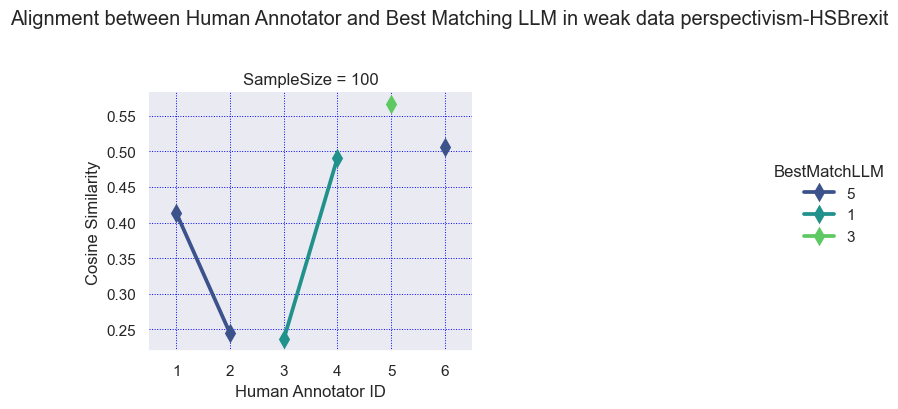}
        
        \vspace{1em} % Maintain space between images and table
        
        {\tiny % Keep tiny font for the table
        \begin{tabular}{p{0.17\linewidth} p{0.37\linewidth} p{0.38\linewidth}} % Use p{} columns for wrapping text
            \toprule
            Human/LLM ID & Persona (strong) & Persona (weak)\\
            \midrule
            1 & Male Muslim Migrant & first generation immigrants from developing countries\\
            2 & Female Muslim Migrant & second generation immigrants and muslim students with Islamic background from developing countries\\
            3 & Neutral foreigner in the UK & migrants\\
            4 & Native English man & researchers with western background, having experience in linguistic annotation\\
            5 & Native English Woman & researchers with western background with NO islamic background\\
            6 & Liberal English person & local that is someone whose ancestors were born in Europe or United Kingdom\\
            \bottomrule % Add bottomrule for professional look
        \end{tabular}
        }
    \end{mdframed}
    \caption{\small Figure showing Prototypical LLM annotators and Alignment with Human Annotators in HSBrexit}
    \label{fig:alignmenthsb}
\end{figure*}

\textbf{Ablation 2}: Figures \ref{fig:alignmentconv} and \ref{fig:alignmenthsb} present an alignment analysis between LLM personas and human annotators. We compute cosine similarity between their annotation vectors. Using sample sizes of 5, 10, 50, and 100, stronger alignment was observed at sizes 50 and 100. In the ConvAbuse \textit{strong} variant, ANN(2–8) showed varying degrees of alignment with LLM Persona 1 (Olivia, female, white, British), while ANN(1) aligns more closely with LLM Persona 4 (Sophia, female, white, from the UK). Other LLM personas (2, 3, 5–8) exhibit no correspondence with any human annotator. We further trained annotator models on annotations from LLM Personas 1 and 4, and evaluated them against human-labeled data. These models showed improved performance, approaching human-level results for both Composite Embedding and SBERT, as shown in Table~\ref{tab:table_align}.

\begin{table*}[t]
\small
\centering
\begin{tabular}{lcccc} \toprule
\textbf{Model} & \textbf{SBERT} & \textbf{User Token} & \textbf{Composite Embedding} & \textbf{Composite Embedding + User Token} \\ \midrule
HL & 85.9 & 88.5 & 85.8 & 88.6 \\ %\hline
LLM & 85.7 & 84.4 & 84.6 & 84.4 \\ %\hline
LLM-H & 83.1 & 83.4 & 84.5 & 84.2 \\ %\hline
LLM(1,4)-H & 85.4 & 82.6 & 85.1 & 85.9 \\ %\hline
\bottomrule
\end{tabular}
\caption{Model performance based on different training and testing label splits: HL (models trained and tested on Human Labels), LLM (models trained and tested on LLM Labels), LLM-H (models trained on LLM Labels, tested on Human Labels), and LLM(1,4)-H (models trained on the most aligned LLM personas 1 and 4 to human labels, tested on Human Labels).}
\label{tab:table_align}
\end{table*}
In the HS-Brexit dataset, alignment is less consistent. In Figure \ref{fig:alignmenthsb}, we see Persona 1, Male Muslim migrant, belonging to the \textit{target} group mapped to annotators 4 and 5 of the human annotators belonging to the \textit{control} group in the strong variant. Human annotators 1–3 belong to the Muslim or migrant group, while annotators 4–6 belong to the group with Western background, denoted as \textit{locals}. Also, Persona 3 of the migrant group representing "neutral foreigner" shows positive alignment in the \textit{weak} variant to the migrant group in human when "Muslim" was removed. These findings suggest that Llama2 includes prototypical personas capable of partially representing multiple human annotators. However, other defined personas fail to map to any observed human annotation patterns (cf. Appendix \ref{sec:align}).

\section{Discussion and Conclusion}
This work investigates Llama2's capacity to generate disaggregated labels for hate speech and offensiveness datasets using predefined personas, under two perspectivism frameworks: strong (individual) and weak (group) data perspectivism. We examine the quality and alignment of LLM-generated annotations with human-annotated datasets and evaluate downstream performance across existing annotator modeling techniques.

Llama2 annotations consistently exhibited higher inter-annotator agreement (Krippendorff’s alpha ranging 0.55–0.91) than human annotations across both ConvAbuse and HS-Brexit datasets, though agreement decreased at higher temperatures. PDF analysis further indicated that LLM annotations tend to converge around features inherent in the model's underlying corpus, suggesting a divergence from human perspectives. As seen in Figure \ref{fig:pdfhs}, the PDF using the soft label distribution of the abusive class shows human annotations aligning towards the non-abusive class, strong perspectivism aligning more towards the abusive class, and weak perspectivism showing a relatively flat and dispersed distribution depicting high uncertainty.

In terms of performance of annotator modeling methods, LLM annotations shifted model efficacy. While prior work confirmed that annotator models trained on human-annotated datasets with high agreement (e.g., ConvAbuse) performed best with the Composite Embedding + User Token model, and those with low agreement (e.g., HS-Brexit) favored the User Token model, our findings with LLM-generated annotations demonstrate that simpler models, specifically SBERT and Composite Embedding models without explicit annotator information, showed improved results. This shift implies that LLM-generated annotations align more with generalized perspectives and are less suited to highly personalized approaches. Comparing the two perspectivism approaches, strong data perspectivism on ConvAbuse, characterized by overlapping and more personalized features, improved the performance of annotator modeling techniques over its weak counterpart. Conversely, weak perspectivism on HS-Brexit, with its contrasting demographic features in groups, yielded improved performance specifically with SBERT and Composite Embedding models, suggesting that contrasting demographic diversity tends to influence the choice of perspectivism approach and annotator modeling performance in LLMs.

Our ablation studies revealed LLM personas do not directly correspond to human annotators. However, as seen in Figure \ref{fig:alignmentconv}, we identified generalized "prototypical persona features" working as representatives of groups of humans (e.g., ANN 2-8 mapping to LLM Persona 1, ANN1 to LLM Persona 4). Swapping the labels of corresponding annotators in the original dataset with these prototypical annotator labels, and evaluating with the human test set, slightly improved results, as seen in Table \ref{tab:table_align}, presenting a novel approach for modeling perspectivism in LLMs. These findings suggest that while LLMs offer insights into subjective domains, their capacity to fully embody external personas remains limited to their underlying corpus, supporting an aggregated view rather than personalization. Future work should focus on standardization and generate more diversified personas, systematically varying features, and expanding evaluation to other LLMs to fully investigate these prototypical attributes and their potential in capturing a wider scope of perspectives.

%Table~\ref{tab:model} compares the performance of the annotator modeling techniques based on human and Llama2-13B annotations, based on strong and weak perspectivism

\section{Limitations}
This study is based on two datasets and focuses exclusively on binary classification tasks for hate and offensive speech detection. One potential limitation is that the data used to train Llama2-13B may have been filtered, reducing its sensitivity to detecting abusive content, potentially influencing the observed results. Our analysis is also limited to this model, and we did not investigate how newer variants of Llama or other LLMs, like GPT 4o, might influence the results.
 The personas used for generating annotations were limited to the demographic features explicitly provided in the original datasets, with slight modifications to fit the perspectivist spectrum. Furthermore, we did not quantify the extent to which the model’s attention was distributed between the persona and the input sentences. Understanding this balance could provide deeper insight into how strongly LLMs personalize their annotations. 
 
 Another limitation of this study arises from the design of the annotation prompt for the HS-Brexit dataset variant, which focused on `hate speech towards Brexit'. However, the prompt was structured to provide general contextual information about Brexit and simulate the prior knowledge of human annotators. A follow-up experiment analysing the model’s attention mechanism revealed that the instance of "Brexit" appearing first in the prompt received a significantly higher attention score of 0.0654 than the "Brexit" label target which received an attention score of 0.0050. Furthermore, when `immigrants' was targeted instead, it received an attention score of 0.0117, which was higher than that given to 'Brexit' as a target. This suggests that the models have learned to recognise plausible targets for hate speech, which warrants further investigation. However, this paper's specific focus is to investigate the impact of Annotator Personas on LLM behaviour across the perspectivism spectrum. It therefore does not include a deep analysis of the model's sensitivity to target plausibility. Nevertheless, we present this as a compelling avenue for future research, while maintaining that our core findings regarding persona-driven perspectivism remain valid within the described experimental setup.  Our codes are publicly available\footnote{\url{https://doi.org/10.5281/zenodo.16744588}} to support future work.

\label{sec:bibtex}

\section*{Acknowledgments}
This research was supported by funding from Hessian.AI. Any opinions, findings, conclusions, or recommendations in this material are those of the authors and do not necessarily reflect the views of Hessian.AI.

% Bibliography entries for the entire Anthology, followed by custom entries
%\bibliography{anthology,custom}
% Custom bibliography entries only
%\bibliography{custom}
\bibliography{rebib}

\clearpage
\appendix

\section*{Appendix}
\label{sec:appendix}

\section{Prompt Design}
\label{sec:prompt}
\textbf{Prompt for Strong HS-Brexit}
\textit{annotate this statement regarding Brexit. As a (persona) in the UK, you must assign a label to the statement. Use 'O' if the statement does not express hate towards Brexit, and '1' if it does express hate towards Brexit. Provide only the number as your response without any additional text or explanation} 
\textit{for example "\#\#\#Annotator:0" or "\#\#\#Annotator:1"}
\\
\\
\textbf{Prompt for Weak HS-Brexit} 
\textit{annotate this statement regarding Brexit. for a group of (persona) in the UK or Europe, you must assign a label to the statement. Use 'O' if the statement does not express hate towards Brexit, and '1' if it does express hate towards Brexit}. \textit{Provide only the number as your response without any additional text or explanation for example "\#\#\# Annotator:0" or "\#\#\#Annotator:1"}
\\
\\
\textbf{Prompt for Weak ConvAbuse}
\textit{annotate these conversations between users and bots. As a group of {persona} within the age range of 19-21, a level 1 English speaker and an Undergraduate student in Gender Studies and Sociology, you must assign a label to the statement. Use 'O' if the statement is not abusive, and '1' if it does express abusiveness by the user towards the bot or agent. Provide only the number as your response without any additional text or explanation. for example "\#\#\# Annotator:0" or "\#\#\# Annotator:]"}
\\
\\
\textbf{Prompt for Strong ConvAbuse}
\textit{annotate these conversations between users and bots. As a {persona} within the age range of 19-21, a level 1 English speaker and an Undergraduate student in Gender Studies and Sociology, you must assign a label to the statement. Use 'O' if the statement is not abusive, and '1' if it does express abusiveness by the user towards the bot or agent. Provide only the number as your response without any additional text or explanation. for example "\#\#\# Annotator:0" or "\#\#\# Annotator:]"}
\\

\section{Persona Descriptions}
\label{sec:persona}
\textbf{HS-Brexit Persona for Strong Perspectives}
\begin{itemize}
    \item Male Muslim Migrant
     \item Female Muslim Migrant
     \item Neutral foreigner in the UK
     \item Native English man
     \item Native English Woman
      \item Liberal English person
\end{itemize}

\textbf{HS-Brexit Persona for Weak Perspectives}
\begin{itemize}
    \item researchers with Western background having experience in linguistic annotation
     \item first or second generation muslim immigrant students from developing countries
\end{itemize}
\textbf{ConvAbuse Persona for Weak Perspectives }
\begin{itemize}
    \item white British female people
     \item white British with non-binary gender orientation
     \item non-binary gender people from the United States
     \item white female people from the United Kingdom
     \item white female from United States
      \item mixed Asian with a non-binary gender orientation
      \item mixed Asian female
      \item white people from the United States with a non-binary gender orientation
\end{itemize}

\textbf{ConvAbuse Persona for Strong Perspectives }
\begin{itemize}
    \item Olivia, a female and white british person
     \item Emma, a female and white british person
     \item Ariel, a white person from the United Kingdom with a non-binary gender orientation
     \item Sophia, a female and white person from the United Kingdom
     \item Katrin, a female and white person from the United Kingdom
      \item Eve, a female and white person from the United Kingdom
      \item a mixed Asian person
      \item a white person from the United States with a non-binary gender orientation
\end{itemize}

\onecolumn
\section{Model performance for Strong and Weak Data Perspectivism}
\label{sec:strong_weak}
\begin{table*}[h]
\small
\centering
\begin{tabular}{lcccccc}
\toprule
\textbf{Model} & $\boldsymbol{\alpha}$ & \textbf{User-Token} & \textbf{Composite} & \textbf{Composite+ User-Token} & \textbf{Multitasking} & \textbf{SBERT} \\
\midrule
\multicolumn{7}{l}{\textbf{Strong Perspectivism}} \\
\midrule
Human          & 0.65                 & 88.5                & 85.8                  & 88.6                  & 82.3                           & 85.9           \\
0              & 0.91                 & 84.1                & 83.0                  & 84.4                  & 46.9                           & 83.1           \\
0.1            & 0.87                 & 84.4                & \textbf{84.6}         & 84.3                  & 81.1                           & \textbf{85.7 }  \\
0.2            & 0.81                 & 80.5                & 81.5                  & 80.0                  & 46.8                           & 81.5           \\
0.5            & 0.68                 & 69.9                & 70.7                  & 71.1                  & 45.1                           & 69.4           \\
0.8            & 0.60                 & 63.5                & 65.1                  & 64.4                  & 62.6                           & 64.6           \\
\midrule
\multicolumn{7}{l}{\textbf{Weak Perspectivism}} \\
\midrule
0              & 0.93                 & 83.7                & 83.7                  & 81.8                  & 69.0                           & 85.2           \\
0.1            & 0.88                 & 80.1                & 79.3                  & 78.3                  & 79.8                           & 82.0           \\
0.2            & 0.82                 & 81.2                & 81.5                  & 81.2                  & 70.3                           & 82.1           \\
0.5            & 0.67                 & 71.7                & 69.7                  & 71.4                  & 64.7                           & 69.5           \\
0.8            & 0.62                 & 61.7                & 61.4                  & 62.3                  & 58.1                           & 61.4           \\
\bottomrule
\end{tabular}
\caption{Performance of Annotator modeling methods for Strong and Weak data Perspectivism \textbf{(ConvAbuse dataset)} across various temperatures.}
\label{tab:strong_weak_convabuse}
\end{table*}

\begin{table*}[h]
\small
\centering
\begin{tabular}{lcccccc}
\toprule
\textbf{Model} & $\boldsymbol{\alpha}$ & \textbf{User-Token} & \textbf{Composite} & \textbf{Composite+ User-Token} & \textbf{Multitasking} & \textbf{SBERT} \\
\midrule
\multicolumn{7}{l}{\textbf{Strong Perspectivism}} \\
\midrule
Human & 0.35 & 77.6 & 67.6 & 77.3 & 71.7 & 68.6 \\
0 & 0.81 & 69.3 & 71.3 & 71.2 & 65.1 & 72.2 \\
0.1 & 0.73 & 69.4 & 71.8 & 71.0 & 61.8 & 69.2 \\
0.2 & 0.67 & 66.3 & 63.8 & 61.9 & 61.4 & 67.2 \\
0.5 & 0.62 & 61.5 & 61.3 & 61.4 & 49.5 & 62.2 \\
0.8 & 0.58 & 52.4 & 56.1 & 54.2 & 51.2 & 56.6 \\
\midrule
\multicolumn{7}{l}{\textbf{Weak Perspectivism}} \\
\midrule
0 & 0.75 & 72.2 & \textbf{72.4} & 71.7 & 60.3 & \textbf{73.2} \\
0.1 & 0.69 & 66.6 & 65.8 & 65.5 & 62.0 & 69.1 \\
0.2 & 0.62 & 62.2 & 63.8 & 69.9 & 59.2 & 66.8 \\
0.5 & 0.54 & 58.0 & 58.4 & 57.9 & 39.2 & 56.1 \\
0.8 & 0.55 & 55.2 & 57.8 & 56.7 & 55.4 & 56.6 \\
\bottomrule
\end{tabular}
\caption{Performance of Annotator modeling methods for Strong and Weak data Perspectivism \textbf{(HS-Brexit dataset)} across various temperatures.}
\label{tab:strong_weak_hsb}
\end{table*}
\vspace{0.5em}

\onecolumn
\section{Probability Density Function for Uncertainty and Annotation Quality}
\begin{figure*}[h]
    \centering
    \includegraphics[scale=0.8]{ 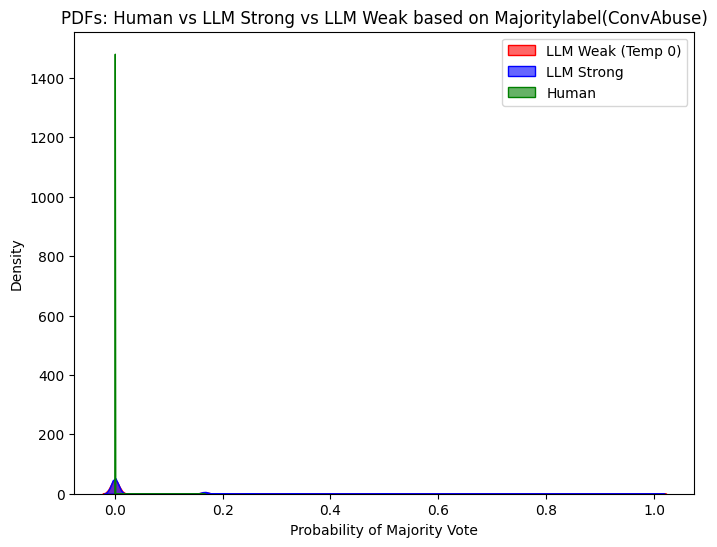}
    \caption{Probability Density Function ConvAbuse Dataset}
    \label{fig:pdf_conv}
\end{figure*}

The Figure \ref {fig:pdf_conv} shows the probability density function of the weak data perspectivism in ConvAbuse using the majority class as a reference point.

\label{sec:pdf}

\onecolumn
\section{Human Vs Persona Alignment and Prototypes ConvAbuse Dataset}
\label{sec:align}
\begin{table}[H]
\centering
\begin{tabular}{lcccr}
\toprule
\textbf{Human Annotator} & \textbf{SampleSize} & \textbf{Best Match LLM Persona} & \textbf{Similarity score} \\
\midrule
2                   & 100                 & 1                     & 0.791            \\
3                   & 100                 & 1                     & 0.894             \\
7                   & 100                 & 1                     & 0.913             \\
8                   & 100                 & 1                     & 0.671             \\
1                   & 100                 & 4                     & 0.707             \\
6                   & 100                 & 1                     & 0.707             \\
4                   & 100                 & 1                     & 0.816             \\
5                   & 100                 & 1                     & 0.745             \\
\bottomrule
\end{tabular}
\caption{Mapping of Human Annotators to Best Matching LLM Personas based on Cosine Similarity.}
\label{tab:annotator_llm_mapping}
\end{table}

\vspace{0.5em}

\textbf{Prototypical Annotators and their Alignment with Human Annotators Across Varying Sample sizes}

\begin{figure*}[h]
    \centering
    \includegraphics[scale=0.6]{ 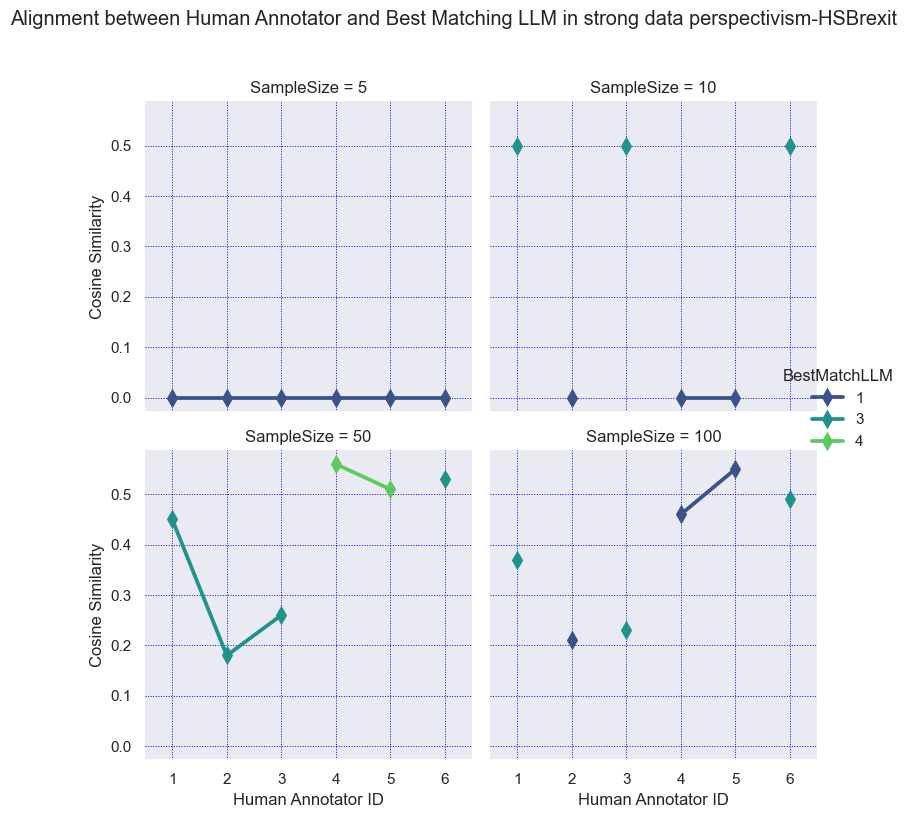}
    \label{fig:alignment_stronghs}
    \caption{Showing the identified Prototypical annotators in HS-Brexit dataset and the alignment with human annotators}
\end{figure*}
\vspace{-3.0em}
\begin{figure*}[h]
    \noindent
    \includegraphics[scale=0.6]{ 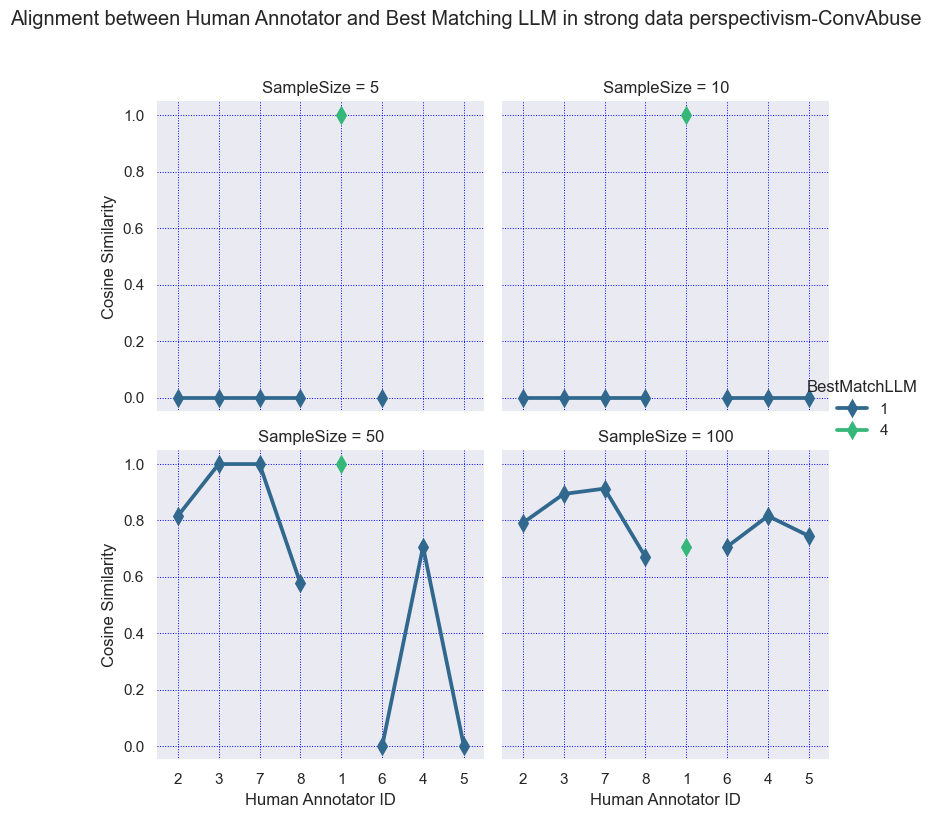}
    \label{fig:alignment_stronghs}
    \caption{Showing the identified Prototypical annotators in ConvAbuse dataset and the alignment with human annotators}
\end{figure*}

\end{document}